\documentclass{article}
\usepackage[a4paper, margin=3cm]{geometry}

\usepackage{url}

\usepackage{authblk}
\usepackage{graphicx}
\usepackage{booktabs}
\usepackage[caption=false]{subfig}
\usepackage[export]{adjustbox}
\usepackage{amsmath, amssymb, amsthm}
\usepackage{enumitem}
\usepackage{hyperref}

\usepackage[numbers]{natbib}

\usepackage[noend,algo2e,ruled,vlined]{algorithm2e} 
\usepackage{algorithm}

\newcommand{\builder}{\text{QXG-Builder}}
\newcommand{\qxg}{\text{QXG}}
\newcommand{\lidar}{LiDAR}

\begin{document}
\title{Towards Trustworthy Automated Driving through Qualitative Scene Understanding and Explanations}
\author[1]{Nassim Belmecheri}
\author[1]{Arnaud Gotlieb}
\author[2]{Nadjib Lazaar}
\author[1]{Helge Spieker}
\affil[1]{Simula Research Laboratory, Oslo, Norway\\\{nassim,arnaud,helge\}@simula.no}
\affil[2]{LIRMM, University of Montpellier, CNRS, Montpellier, France\\nadjib.lazaar@lirmm.fr}
\date{}

\maketitle

\begin{abstract}
Understanding driving scenes and communicating automated vehicle decisions are key requirements for trustworthy automated driving. In this article, we introduce the Qualitative Explainable Graph (QXG), which is a unified symbolic and qualitative representation for scene understanding in urban mobility. The QXG enables interpreting an automated vehicle's environment using sensor data and machine learning models. It utilizes spatio-temporal graphs and qualitative constraints to extract scene semantics from raw sensor inputs, such as LiDAR and camera data, offering an interpretable scene model. A QXG can be incrementally constructed in real-time, making it a versatile tool for in-vehicle explanations across various sensor types. Our research showcases the potential of QXG, particularly in the context of automated driving, where it can rationalize decisions by linking the graph with observed actions. These explanations can serve diverse purposes, from informing passengers and alerting vulnerable road users to enabling post-hoc analysis of prior behaviors.

\end{abstract}

\section{Introduction}

Artificial Intelligence (AI) methods nowadays are at the center of automated driving and connected mobility, including perception and scene understanding~\cite{yang_scene_2019,muhammad_vision-based_2022,muhammad_deep_2021}. However, passing control to an AI-based system and trusting its decisions requires the ability to request explanations for these decisions~\cite{omeiza_explanations_2022}. Societal acceptance of automated driving significantly depends on these AI models' trustworthiness, transparency, and reliability~\cite{nastjuk_what_2020}. 
Still, this is an open challenge, as many of the state-of-the-art machine learning (ML) models are opaque and not inherently explainable by themselves~\cite{atakishiyev_explainable_2023}.

In recent years, several explainable AI methods with a focus on automated driving have been proposed. Following \cite{atakishiyev_explainable_2023}, they fall into three main categories: 
a) \emph{Vision-based} explainable AI related to highlighting the area of an image that influences a perception model towards a certain output \cite{omeiza_explanations_2022}; 
b) \emph{Feature-based importance scores} quantify the influence of each input feature on the model output; and 
c) \emph{Textual-based} explainable AI that aims to formulate explanations as intelligible arguments using natural language processing \cite{kim_toward_2021}.
Unfortunately, automated support for multi-sensor and video-based scene explanation is still restricted to quantitative analysis, e.g., saliency heatmaps \cite{omeiza_explanations_2022}.

In this work, we exploit qualitative methods for scene understanding by using \emph{Qualitative Explainable Graphs (QXG)} and, based on this representation, we propose a method for action explanation through simple classification models. A QXG captures the spatio-temporal dynamics of a scene via qualitative algebras, i.e., a description of the relative positions (e.g., pedestrian north of ego car), a qualitative distance (e.g., pedestrian far from ego car) and their direction towards each other (e.g., ego car approaching static pedestrian). From these graphs, ML models are trained to explain actions and their causes. 
Our results on the real-world nuScenes dataset \cite{caesar_nuscenes_2020} show that the QXG can be efficiently constructed incrementally in real-time and is capable of correctly explaining actions.

\section{Background \& Related Work}

\paragraph{Qualitative Calculi.}

A \emph{qualitative calculus} (QC) is a computational method designed to analyze and understand qualitative relationships among physical attributes, such as position, velocity, and acceleration, without reliance on precise quantitative data~\cite{dylla_survey_2015}. 

This approach of computation can be perceived through different qualitative algebras, that can be used for modeling temporal dynamics, spatial relationships, or even a combination of both
\cite{allen_maintaining_1983,renz_qualitative_2007}.

In the field of automated driving, qualitative reasoning plays an important role. It can be utilized through the adoption of ontologies \cite{westhofen_using_2022} and neurosymbolic online abduction \cite{suchan_commonsense_2021}. These methods enable the encoding of driving scenarios and the complex dynamics of traffic, especially when obtaining precise measurements presents challenges or is completely unfeasible.

QC have broad application in spatio-temporal reasoning, where they simplify the description of relationships between sets of objects in both space and time. 
For example, QC can describe the positioning, distance, or orientation of vulnerable road users w.r.t. a vehicle in abstract, understandable terms, e.g., the pedestrian is located northwest and far from the vehicle, rather than giving only the absolute coordinates of each object.
This is relevant in the context of automated driving, where understanding the spatial and temporal interactions of various entities is crucial for safe and efficient operations.

In this work, we rely on four complementary qualitative calculi \cite{dylla_survey_2015} for all spatial aspects. Other algebras may be considered as well, but our analysis and experiments demonstrate the combination of these four ones is appropriate and sufficient:

\begin{enumerate}[leftmargin=*]
\item \textbf{Qualitative Distance Calculus (QDC)} \cite{renz_qualitative_2007}: This calculus qualitatively specifies the distances between objects, i.e., determining whether two objects are close or far from each other. It does so without requiring precise distance measurements, making it a valuable tool for spatial reasoning. In the context of automated driving, it helps in expressing distance-based relationships between objects without requiring fine-grained measurements.

\item \textbf{Rectangle Algebra (RA)} \cite{renz_qualitative_2007}: The Rectangle Algebra describes relative object positioning and considers that objects have a dimensionality and are not only points. 
In this calculus, each object is represented by a rectangle, e.g., taken from the bounding box of the sensor detections. 
Relations between the objects are constructed from the 13 relations of Allen's interval algebra \cite{allen_maintaining_1983} in both the x- and y-axis of the internal coordinate system. 
By using RA, we can better understand how objects are positioned about each other and their spatial orientations.

\item \textbf{Basic Qualitative Trajectory Calculus (QTC)} \cite{dylla_survey_2015}:
This calculus describes the orientation of objects towards each other. 
It allows for reasoning about the motion and paths of objects without the need for precise numerical data. 
Unlike the other calculi, QTC inherently involves a temporal aspect, since it must consider the direction of movement between two frames.

\item \textbf{Star Calculus (STAR)} \cite{renz_qualitative_2007}:
The Star Calculus serves to represent and reason about spatial in the Euclidean space. It is particularly useful for describing regions of influence, zones, and coverage areas in different driving scenarios. In our work, we specifically apply the STAR$_4$ variant, which divides the surrounding space into four quarters. By utilizing the Star Calculus, we can gain insights into spatial regions and their relationships.
\end{enumerate}

\paragraph{Qualitative Scene Understanding.}
Scene understanding is a concept that involves collecting, organizing and analyzing spatial and temporal information related to various objects, including vehicles, vulnerable road users, and stationary elements, e.g., traffic lights, cones, or barriers. 
This analysis is applied to a sequence of frames, with the purpose of comprehending the dynamics in scene \cite{xue_survey_2018}. 
The groundwork of scene understanding involves perception-related tasks, such as object detection and image segmentation, which are essential for recognizing and identifying objects within an observed environment \cite{muhammad_vision-based_2022}. Scene understanding is then performed on a higher level, where we extract and analyze the qualitative description of the scene through the aforementioned QC.

Qualitative scene understanding is performed one abstraction level above the quantitative analysis by capturing the objects and their spatio-temporal relationships beyond a coarse discretization of the quantitative information.
It does not replace quantitative scene analysis and representations, which are used for many automated driving tasks, but it is a parallel representation of the scene with complementary properties.
Qualitative reasoning makes use of qualitative calculi and allow to check the consistency of the relations, which can highlight wrong or noisy sensor data, and derive missing information through composition, within a formally defined framework.
A qualitative representation can thereby enhance existing representations by being an interpretable and explainable additional layer.

In the context of automated driving, we formally define a scene as a sequence of
$n$ frames, represented as $\mathcal{S}=\langle f_1,\ldots, f_n \rangle$. 
One upstream component for scene understanding is object detection and tracking.
This process requires identifying objects in a given frame, denoted as $f_k$ within $\mathcal{S}$, determining their respective bounding boxes, and monitoring their movements concerning previously detected objects in earlier frames. In $\mathcal{S}$, we detect a set of $m$ objects, collectively denoted as $\mathcal{O}=\{ o_1,\ldots o_m\}$. Each object $o_i$ is observed to appear in at least one frame $f_j\in \mathcal{S}$, providing a temporal dimension to the understanding of the scene.

Scene understanding primarily concerns the assessment of the situational context, where the goal is to comprehend the evolving relationships between objects, the dynamics of the scene, and how the various elements interact. 
Many applications of scene understanding can be imagined in the context of automated driving, here are some important ones: 

\begin{itemize}
    \item \textbf{Supporting Decision-Making in automated driving vehicles:} 
    A clear understanding of the environment through qualitative scene understanding can play an important role in supporting automated driving systems to make decisions, especially in tasks like path planning, obstacle avoidance, etc. Having a qualitative representation can help to understand the surroundings, thus forming a basis for safe and efficient automated navigation, complementary to the main decision-making policy.

\item \textbf{Precise Object Trajectory Prediction:} Trajectory prediction is a crucial automated navigation task, the goal here is to predict the object/entity trajectories. Having an insight into the dynamics of objects within the scene, including vehicles, cyclists and pedestrians, scene understanding supports predictive models with the essential information needed to help forecast the future paths and behaviors of these entities through an abstract qualitative representation and the consistency it presents.

\item \textbf{Transparent Explanations and In-Depth Analyses:} 
Scene understanding introduces an advanced mechanism for generating transparent and intuitive explanations of actions and decisions executed within a monitored environment. This is important to show trustworthiness to human stakeholders. Moreover, it serves as an instrumental tool for both real-time and in-depth analysis. The generated explanations can function as confirmations or alerts for vulnerable road users and other road users in safety-critical situations. This approach aims to increase the interoperability and reliability of the automated vehicle decision in risky scenarios.
\end{itemize}

\section{Qualitative Explainable Graph}
\begin{figure}[t]
    \centering
    \includegraphics[scale=0.3,width=\textwidth]{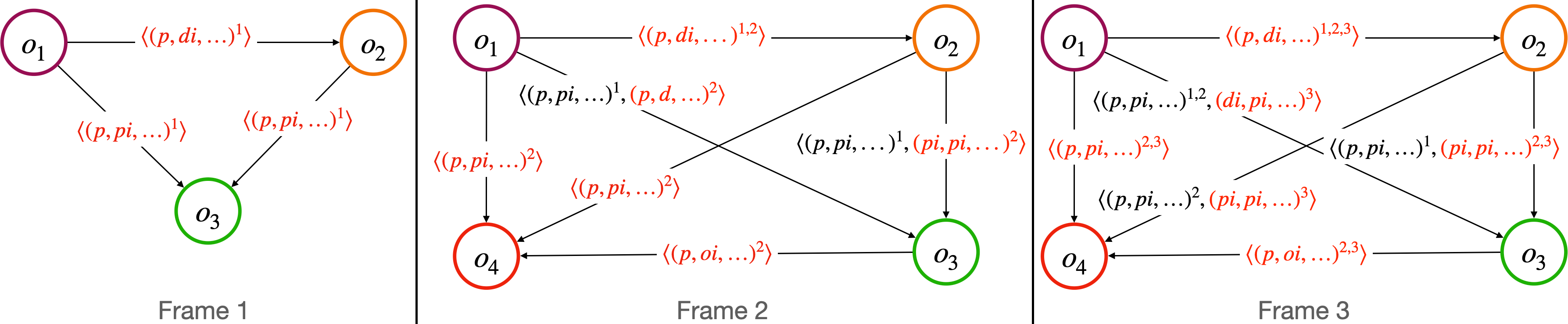}
    \caption{Illustration of the incremental construction of the QXG over three frames. For the sake of simplicity, only the rectangle algebra (RA) relation is depicted on the edges.}
    \label{fig:qxgbuilder}
\end{figure}

In this article, we introduce the Qualitative Explainable Graph (QXG) serving as a scene representation based on qualitative relations between the various scene entities. The QXG representation is parameterized by multiple qualitative calculi that capture different aspects of the spatial and temporal relations in a scene. The QXG captures the complex relations between object pairs throughout the temporal evolution of a scene, all supported by spatial relations and their change over time. The choice of the spatial calculi is a configurable parameter of the QXG. 
This adaptability allows us to adapt to varying needs in terms of granularity and specific use cases. 
Figure~\ref{fig:qxgbuilder} illustrates the incremental construction of the QXG associated with three successive frames showing interactions between four objects. The relations on the edges are updated at each frame and capture the qualitative evolution of the scene.

For the sake of simplicity, we use four distinct calculi to effectively capture the necessary spatial relations within a scene. 
Each of these calculi contributes a unique perspective on spatial relations, and their union ensures coverage of the diverse aspects for understanding and explaining scenes via qualitative graphs. However, we must emphasize that the formulation of the QXG and its utility extends beyond the specific choice of calculi. As long as the selected calculi are provided with the expressive capacity to cover at least the relative positioning and distance between objects, the QXG remains highly adaptable. 
Moreover, for complex use cases with the necessity of high-fidelity, but still interpretable representations, the addition of other qualitative calculi \cite{dylla_survey_2015} can be considered, further enriching the QXG representation. 
Overall, we claim that the QXG is an intuitive and easy-to-understand way of representing scenes.

\SetKwInOut{InOutput}{In Out}
\SetKwInOut{Input}{In}
\SetKwInOut{Output}{Out}

 \subsection{Qualitative Explainable Graph Builder}\label{sec:algorithm}
In this section, we present \builder{} (see Algorithm~\ref{alg:builder}), an algorithm introduced for the construction of the QXG to represent scenes through constraint acquisition. {\it Constraint Acquisition} is a symbolic ML approach that represents learned concepts in the form of constraint networks \cite{bessiere_constraint_2017}. This process involves the classification of positive and negative instances of the target concept by an oracle, which can be a human or an automated procedure. In our work, where the constraint network comprises qualitative constraints, we use an existing constraint acquisition process named GEQCA (GEneric Qualitative Constraint Acquisition). GEQCA is an active learning process rooted in qualitative reasoning that can be used for acquiring constraints between pairs of entities \cite{belaid_geqca_2022}.

The QXG of a scene $\mathcal{S}$ is formally defined as a pair $(\mathcal{O}, V)$, wherein $\mathcal{O}$ represents the set of objects, and $V$ the labeled edges that include the qualitative relations between objects across the sequence of frames. \builder{} undertakes an iterative process across all frames (main loop at line~\ref{qxg:loop1}). 
For a given frame $f_k$, \builder{} starts by extracting the objects present in $f_k$ and tracking those persisting from previous frames. This is executed using the \texttt{objectDT} function at line~\ref{qxg:DT}, where each detected object is enclosed by a bounding box, capturing the (almost) smallest surrounding box of the entity (e.g., car, pedestrian, cycle). Then, \builder{} iterates through all pairs of objects to acquire constraints between them using GEQCA (line~\ref{qxg:geqca}). In our context, the frame $f_k$ serves as an oracle, validating or rejecting relations between a pair of objects. The output of GEQCA for a pair of objects $(o_i,o_j)$ in a frame $f_k$ is denoted by $V_{ij}[k]$, expressing the atomic relations derived from various qualitative algebras, representing the connection between the bounding boxes of $o_i$ and $o_j$ at frame $f_k \in \mathcal{S}$. 
Through this incremental process, the QXG evolves by updating objects in $\mathcal{O}$ and the relations between all pairs of objects frame by frame in $V$.

{

\begin{algorithm}

	\LinesNumbered
\setcounter{AlgoLine}{0}
	\SetAlgoLined
	\Input{Sequence of frames $\mathcal{S}=\langle f_1,\ldots,f_n\rangle$; {\tt \ \ // Scene defined as a sequence of frames}\\
         }
	\Output{Qualitative Explainable Graph $\qxg=(\mathcal{O},{V})$; {\tt \ \ // $\mathcal{O}$:objects, $V$:labeled edges} }
\BlankLine

\For{$k\in 1..n$}{ \label{qxg:loop1}
$\mathcal{O} \gets \texttt{objectDT}(\mathcal{S},f_k)$; {\tt \ \ \ \ \ \ \ \ \ \ // Incremental Object Detection and Tracking} \label{qxg:DT}

\ForEach{$(o_i,o_j)\in \mathcal{O}^2: i<j$}	{ \label{qxg:loop2}
					
			 $V_{ij}[k] \gets \text{\sc GEQCA}(f_k,o_i,o_j)$;  {\tt \ \ // GEneric Qualitative Constraint Acquisition} \label{qxg:geqca}

					}

}

\Return $(\mathcal{O}, {V})$;

	\caption{\builder}\label{alg:builder}
	
\end{algorithm}

}

Even though it is not in the scope of the discussion in this paper, the algorithm \builder{} can be easily adapted for a stream-based, incremental construction of the QXG during real-world deployment. 
Such revision needs to take the input scene as a streaming data source, allowing for QXG updates with each incoming frame. 
Additionally, it is important to highlight that for practical applications integrated with a real AD system, defining a time window is a straightforward way to ensure that the QXG maintains a representation of the $k$ most recent incoming frames.

\section{Bridging the Gap: Connecting Actions and QXG}

The QXG serves as a descriptive framework for scenes, showcasing the spatio-temporal relationships among objects. When the goal is to explain events within a scene, it becomes crucial to establish connections between the abstract observations covered in the QXG and the actions that precipitate specific occurrences. Importantly, these actions can not only originate from the ego vehicle; but can also come from any object within the scene, be it road users, other vehicles, or lifeless elements (e.g., a rolling ball). This flexibility arises from the QXG's capability to capture relations among all objects.

\subsection{Action Description}
In automated driving systems, actions are many-sided, covering a range of vehicle maneuvers. These maneuvers can be broadly classified as follows:
\begin{itemize}
    \setlength{\itemsep}{0.3em}
    \setlength{\parskip}{-0.3em}
    \setlength{\parsep}{-0.3em}

\item The steering operations involve actions that influence the vehicle's direction. For instance, adjustments in steering angles guide the vehicle's ability in traversing different road types and executing turns \cite{campbell_autonomous_2010,montemerlo_junior_2009}. Steering plays a role in lane management for the vehicle to maintain its position within a lane. This includes lane changes and ensuring safe traffic flow \cite{shladover_review_1995}

\item By managing acceleration and deceleration, the vehicle can adapt its speed depending on the dynamics of the environment, ensuring smooth transitions and safe interactions with other objects/entities on the road \cite{thrun_toward_2010}.

\item In obstacle avoidance, actions are executed to avoid collisions with obstacles or other vehicles. These actions contribute to the vehicle's ability to respond dynamically to unexpected obstacles or situations on the road \cite{dolgov_path_2010}.

\end{itemize}

\subsection{Action Extraction}
The \textit{Action Extraction} phase prepares the creation of a labeled dataset that, in turn, can be used as a basis for action explanation and training of action explainability classifiers. We apply two heuristics for action extraction:

\begin{itemize}
    \item[(i)] \textbf{Analyzing Movement Patterns:} One well-known approach is based on the analysis of the movement patterns taken by the ego car. This is precisely accomplished through trajectory analysis, a simple tracking of the car's position over time. This will help in identifying the ego car's dynamic.

\item[(ii)] \textbf{Thresholding Methods:} Another effective technique is based on applying thresholds on key parameters such as speed, acceleration, and angular velocity \cite{ruoyu_data-driven_2020}. For instance, a sudden increase in speed may signify an ``acceleration'' action, while a significant change in angular velocity could signal a ``turn''. These threshold-based methods are approximate, however they help identify various actions based on quantitative changes in these parameters.

\end{itemize}

In our approach, the actions extracted through these processes fall into two distinct categories:

\begin{itemize}
\item \textbf{Steering and Turning Maneuvers:} This category considers the change in the vehicle's direction concerning angular velocity. It provides information about the adjustments made to the vehicle's trajectory. For turning, actions are further classified as ``Left'', ``Right'', or ``No Turn'' based on the magnitude of the change in angular velocity.

\item \textbf{Acceleration and Deceleration:} Actions in this realm are categorized as ``Accelerate'', ``Decelerate'', or ``Maintain Speed (Cruising)'' based on the observed velocity changes.
\end{itemize}

By having an automatic process for actions categorization, we establish a ground truth for validating our approach, particularly in creating labeled datasets essential for training and explaining the actions of the automated vehicle.

\subsection{Explaining Actions through QXG}
We will now detail how the QXG is used to explain actions.
The goal is to highlight an object pair and its most recent chain of relations, i.e., how their spatio-temporal relationship evolved, that is most likely to have caused an observed action.
A key component of this process involves an action explainability classifier, that is trained on historical data and can score how commonly a particular relation chain is observed in the context of an action.
It is important to note, that the QXG-based explanation process is not dependent on a specific object in the graph, e.g., an ego car, but can be applied to any action taken by any object in the graph, making it a generalized explanation method for scene understanding.

\begin{figure}[t]
    \centering
    \includegraphics[width=0.65\textwidth]{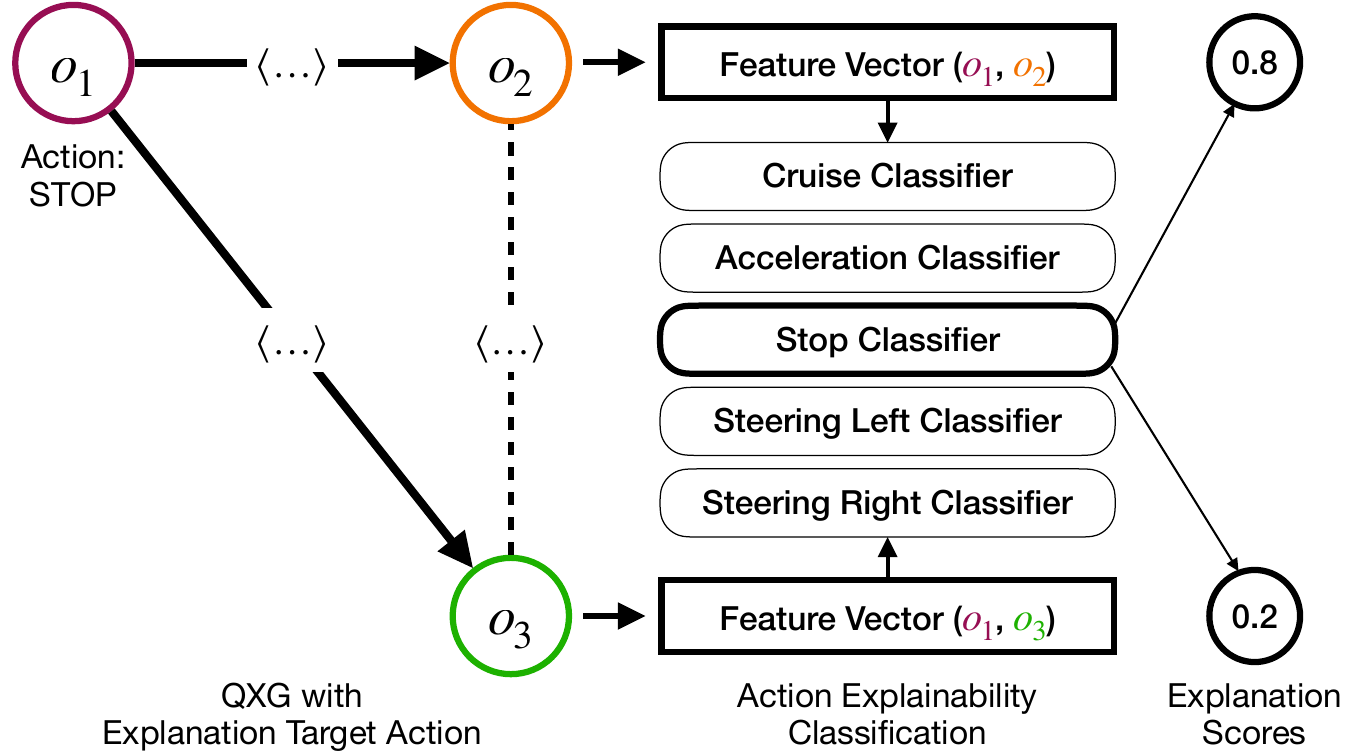}
   \caption{QXG Action Explanation Process: Previous relations extracted from the action-annotated QXG are classified based on the target action.\label{fig:actionexplanation}}
\end{figure}

We show an example for the QXG-based explanation process in Figure~\ref{fig:actionexplanation}.
Here, the relevant object pairs to explain the Stop action by $o_1$ are $(o_1,o_2)$ and $(o_1,o_3)$.
The relation chains for each of these pairs are transformed into the corresponding feature vector and passed to the corresponding classifier model, i.e. the stop classifier in our example.
The classifier assigns scores to these object-relation chains, reflecting how well they align with the action to be explained. Object pairs receiving high explanation scores are identified as representative explanations, offering a rationalization for the observed action, i.e. in our example $(o_1,o_2)$ with the score of 0.8 versus $(o_1,o_3)$ with 0.2. 
Conversely, object pairs receiving low explanation scores, or high scores for alternative actions, warrant scrutiny, as they may signify potential trade-offs or conflicting information in the preceding decision-making stages.
Since we observe the scene from an outsider's point of view, we consider QXG-based explanations to be post-hoc rationalizations, that is, they justify why an action might have been chosen without internal knowledge of the actor.

More precisely, the method of QXG-based action explanations consists of the following steps:
\begin{enumerate}
    \item \textbf{Action Labeling for Scenes:} The initial step involves labeling the QXG with actions as a preparation for the training of the action explainability classifiers. For explanation purposes, only the QXG and the single action to be explained are required. The action labels are expected as an external input, either from an annotated dataset, through observation, or from the control system of an object in the graph. In our scenario, where actions are extracted from observed scenes denoted as $\mathcal{S^*}$, this step is formalized with the extraction function $F: \mathcal{S^*} \rightarrow \mathcal{A}$. This function extracts actions $a_i \in \mathcal{A}$ through trajectory analysis or threshold's methods \cite{ruoyu_data-driven_2020}. For each action $a_i$, object-pair relation chains within the temporal context of the last $n$ frames in the QXG are identified. These chains are translated into feature vectors $x_{a_i}$, forming the basis of our labeled dataset $\mathcal{D} = {(x_{a_i}, a_i)}$. Crucially, each object pair in the scene is associated with the corresponding action $a_i$. 

\item \textbf{Action Classifier Training:} Two types of classifiers can be trained, which take different approaches to explanation scoring. First, individual classifiers $\mathcal{C}_I=\{C_{a_i}: a_i \in \mathcal{A}\}$ are trained for {\it anomaly detection}, where each $C_{a_i}: x_{a_i} \mapsto \{\text{action } a_i,\text{ no action } a_i \}$ aims to discern the anomalous nature of a specific action $a_i$. This is important for real-world datasets, often devoid of abnormal interactions between objects. Anomaly detection classifiers identify unusual interactions and can thus be used to furnish explanations. Alternatively, a multi-class classifier $C_M: x \mapsto \mathcal{A}$ can be trained to predict one of the multiple actions, recognizing specific actions from the feature vectors. Both approaches contribute to recognizing patterns indicative of certain actions.

\item \textbf{Explanation Generation:} Upon observing an action $a_i$, the relevant classifier $C_{a_i}$ evaluates, based on the past $n$ frames, the likelihood of each object-pair interaction $(o_{i},o_{j}) \in \mathcal{O}$ that affirms or rejects the observed action.

\end{enumerate}

The classification approach can provide two perspectives on action explanation:

\begin{itemize}
    \item \textbf{Affirmation of the action}: The classifier $C_{a_i}$ affirms the occurrence of action $a_i$ by identifying object pairs that align with the predicted action. In this role, it serves as an action explanation, providing justification for the decision made.

    \item \textbf{Rejection of the action}: Conversely, $C_{a_i}$ rejects the set of object-pairs $\mathcal{O}$ for anomalies that contradict the expected action $a_i$. This rejection capability allows the system to identify instances where the observed action might not be accurate or might have alternative explanations. 
\end{itemize}

This dual functionality enables the QXG-based action explanations not only to support decision rationalization, i.e., providing support for why an action was taken, but potentially also action rejection, i.e., why an action was not taken.
This process is particularly essential for validating behavior in safety-critical situations. By exploring scenarios where an alternative action might have been more appropriate, we gain a deeper understanding of the decision-making processes in the context of automated driving.
In total, the single QXG representation of a scene allows a more comprehensive understanding and analysis of the scene dynamics.

For the remainder of this article, our primary focus is on providing explanations that explain and rationalize the selection of action $a_i$ by identifying the causing object-pairs $(o_{i},o_{j})$.

\section{Experimental Evaluation}
We evaluated our proposed approach through two main sources: the nuScenes dataset \cite{caesar_nuscenes_2020} and custom scenarios tailored to simulate critical automated driving situations. These critical situations are not always covered by existing open-source datasets like nuScenes.
The limitations of current automated driving datasets, especially the lack of action labels for the ego vehicle, impede a detailed analysis of behavioral patterns. To address this challenge, we drew inspiration from existing works \cite{thrun_toward_2010, campbell_autonomous_2010, shladover_review_1995, montemerlo_junior_2009, burgard_map-based_2008}. We devised a heuristic that utilizes predefined threshold values to infer the actions of objects, placing particular emphasis on the ego vehicle. Importantly, our heuristic is designed with flexibility, allowing for potential extensions to other objects within the dataset.

Our experimental evaluation aims at responding to three research questions:
\begin{itemize} \setlength{\itemsep}{0.3em} \setlength{\parskip}{-0.3em} \setlength{\parsep}{-0.2em}
    \item {\bf (RQ1) Can QXG be constructed in real-time?} There are at least two possible exploitation routes for QXG in scene understanding, either in real-time to interpret actions while they are performed or in a post-hoc manner to interpret actions after their completion;
    \item {\bf (RQ2) Among different available classifiers for action recognition in scene understanding, which one provides the best accuracy results?} Various action classifiers can be trained, and we want to determine which ones have the best accuracy to avoid (as much as possible) misclassified actions;
    \item {\bf (RQ3) In scene understanding, is QXG relevant for action interpretation in safety-critical and real-life scenarios?} We argue in our work that QXG can be exploited for real-time action recognition. It is thus of interest to determine exactly the conditions of QXG exploitation and in particular if the QXG can be used in safety-critical and/or real-life scenarios. 
\end{itemize}

\subsection{Experimental Setup}
For this evaluation, we consider five actions for an ego car: {\em Accelerate}, {\em Stop}, {\em Cruising}, {\em Steering Left}, and {\em Steering Right}. These actions are defined based on the following motion vectors and their associated magnitudes:
\begin{itemize} \setlength{\itemsep}{0.3em} \setlength{\parskip}{-0.3em} \setlength{\parsep}{-0.2em}
    \item Velocity vector, \( \vec{v} \), with magnitude \( ||\vec{v}|| \).
    \item Acceleration vector, \( \vec{a} \), with magnitude \( ||\vec{a}|| \).
    \item Angular velocity vector, \( \vec{\omega} \), with magnitude \( ||\vec{\omega}|| \).
\end{itemize}

\noindent
\textbf{Dataset Labeling.}
The actions are labeled based on the following criteria:
\begin{itemize} \setlength{\itemsep}{0.3em} \setlength{\parskip}{-0.3em} \setlength{\parsep}{-0.2em}
    \item Labeled as {\bf Stop} if the magnitudes of the velocity and the acceleration vectors are below small thresholds (noted  $\epsilon_v$ and $\epsilon_a$): 
    \[ ||\vec{v}|| < \epsilon_v \text{ and } ||\vec{a}|| < \epsilon_a \]
    
    \item Labeled as {\bf Accelerate} if the magnitude of the acceleration vector is strictly positive, and the magnitude of the velocity vector is either greater than a threshold \( \epsilon_v \) or approximately zero, i.e.,
    \[ ||\vec{a}|| > 0 \text{ and } (||\vec{v}|| > \epsilon_v \text{ or } ||\vec{v}|| \approx 0). \]
    
    \item Labeled as {\bf Steering Right} if the magnitude of the acceleration vector is strictly positive, and the magnitude of the angular velocity vector exceeds a positive threshold \( \epsilon_{\omega_{\text{right}}} \), indicating a significant rightward turn, i.e.,
    \[ ||\vec{a}|| > 0 \text{ and } ||\vec{\omega}|| > \epsilon_{\omega_{\text{right}}}. \]
    
    \item Labeled as {\bf Steering Left} if the magnitude of the acceleration vector is strictly positive, and the magnitude of the angular velocity vector is less than a negative threshold \( -\epsilon_{\omega_{\text{left}}} \), indicating a significant leftward turn, i.e.,
    \[ ||\vec{a}|| > 0 \text{ and } ||\vec{\omega}|| < -\epsilon_{\omega_{\text{left}}}. \]
    
    \item Labeled as \textbf{Cruising} if none of the above conditions are met, the vehicle is considered to be in a ``cruising'' state.
\end{itemize}

In our approach, the comprehension of the ego vehicle's action involves analyzing the preceding $n$ frames. Established research \cite{ou_predicting_2018,xu_end--end_2017,chi_learning_2017} underscores the significance of evaluating a window of $3$ to $10$ frames from the past for effectively predicting an impending action. Adhering to this guideline, we opt for a sequence length of $5$ frames. Consequently, we assign a label to each $5$-frame sequence based on the final action observed in the $5^{th}$ frame of that sequence. This labeling strategy ensures a close alignment of each action with the ego vehicle's most recent maneuver.

For all experiments, we rely on the implementation and standard configuration of the classifiers from scikit-learn~\cite{pedregosa_scikit-learn_2011}.

\subsection{(RQ1) QXG Construction}
We evaluate the action explanations using $850$ scenes from the nuScenes dataset~\cite{caesar_nuscenes_2020}. 
On these $850$ scenes, the QXGs are incrementally constructed, frame-by-frame, from the top \lidar{} view.

\begin{figure}[!ht]
\centering
\includegraphics[width=\linewidth]{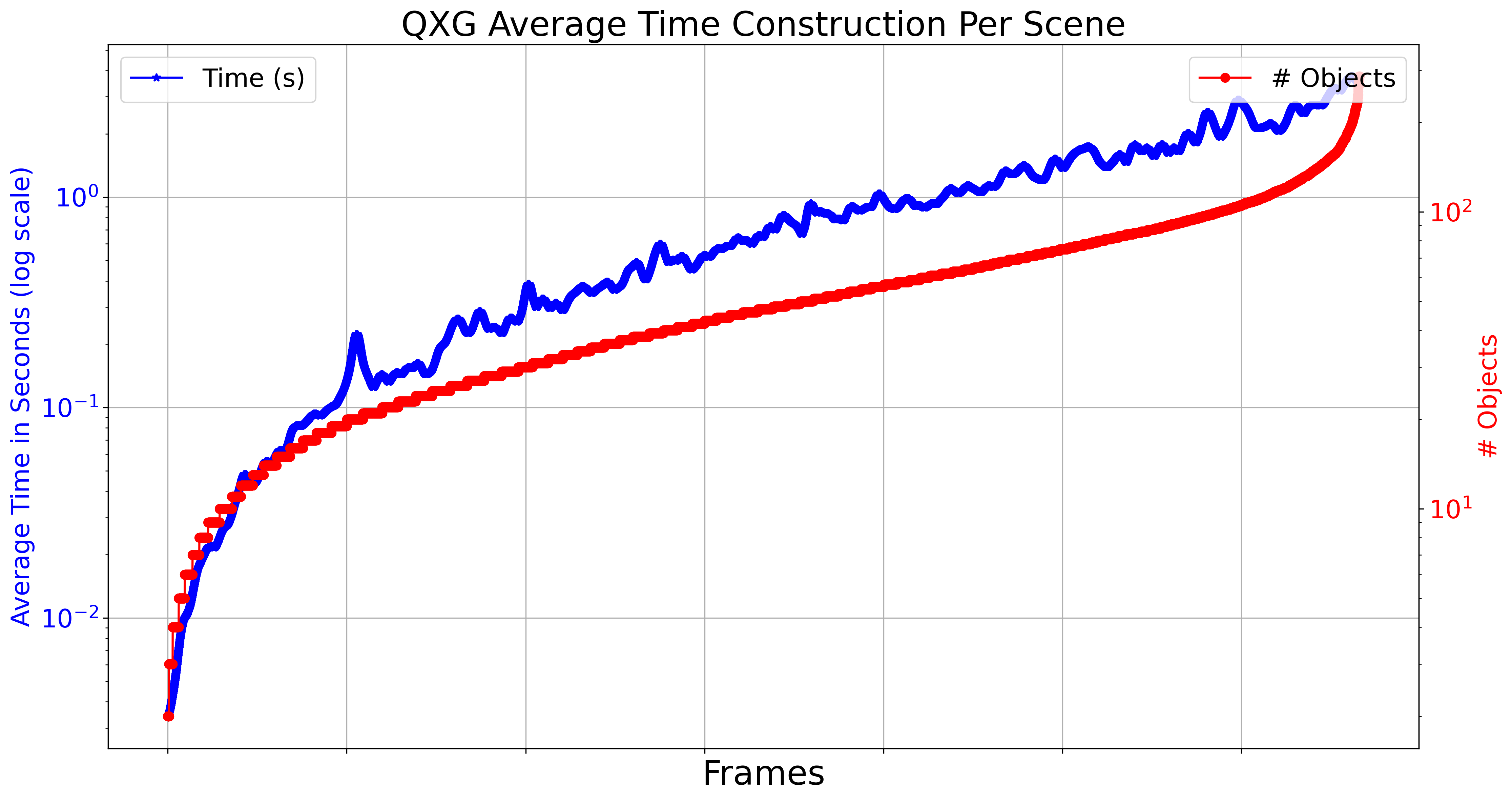}
\caption{Construction time of QXG per frame}
\label{fig:qxg}
\end{figure}

Figure~\ref{fig:qxg} illustrates the time it takes to construct the QXG for each frame in every scene, alongside the total number of objects present. 
The blue line shows the extraction time per frame and the red line the total number of objects per scene. 
The figure shows extraction times consistently below one second for most frames in the dataset, scaling proportionately with the increasing complexity and quantity of objects within the scenes. This indicates that the construction of QXGs can be efficiently performed in real-time until a certain number of objects. We therefore answer RQ1 positively by showing that the construction and maintenance of QXG over a long-term scene can be performed in real-time, provided that the number of objects in the scene does not exceed $50-100$.

\subsection{(RQ2) One-Class vs. Multi-Class Classification Performance}

\subsubsection{Evaluation of One-Class Classifiers}
We have trained multiple one-class classifiers as interpretable action explainability classifiers across $595$ scenes. These one-class classifiers recognize behavior occurring together with their respective action as normal, marking all other behaviors as abnormal outliers.

We evaluate the classifiers on a set of $255$ held-out scenes and report \textit{Recall} and \textit{Precision}, measuring the sensitivity and predictive performance of the classifiers. Recall is the classifier's ability to correctly identify instances of the single positive class, i.e., how many instances were correctly identified as positive. Precision measures how many positive instances from the dataset were classified as positive. We consider four one-class classifiers: Random Isolation Forests, One-Class SVM, Local Outlier Factor, and Elliptic Envelope. Results are shown in Table~\ref{tab:aeresults}.

\begin{table}[ht]
\centering
\caption{One-Class Classifier Results with Recall and Precision Metrics\label{tab:aeresults}}
\begin{adjustbox}{width=\textwidth,center}
\begin{tabular}{lccccccccccc}
\toprule
Classifier & \multicolumn{2}{c}{Cruising} & \multicolumn{2}{c}{Accelerating} & \multicolumn{2}{c}{Stopping} & \multicolumn{2}{c}{Steering Right} & \multicolumn{2}{c}{Steering Left} \\
\cmidrule(r){2-3} \cmidrule(lr){4-5} \cmidrule(lr){6-7} \cmidrule(lr){8-9} \cmidrule(l){10-11}
& Recall & Precision & Recall & Precision & Recall & Precision & Recall & Precision & Recall & Precision \\
\midrule
Random Isolation Forests & 88.8\,\% & 100\,\% & 90.3\,\% & 100\,\% & 87.7\,\% & 100\,\% & 92.0\,\% & 100\,\% & 92.5\,\% & 100\,\% \\
One-Class SVM & 94.4\,\% & 100\,\% & 99.3\,\% & 100\,\% & 38.6\,\% & 100\,\% & 99.4\,\% & 100\,\% & 94.5\,\% & 100\,\% \\
Local Outlier Factor & 69.5\,\% & 100\,\% & 75.1\,\% & 100\,\% & 67.6\,\% & 100\,\% & 75.0\,\% & 100\,\% & 69.7\,\% & 100\,\% \\
Elliptic Envelope & 91.1\,\% & 100\,\% & 91.8\,\% & 100\,\% & 88.4\,\% & 100\,\% & 92.0\,\% & 100\,\% & 92.2\,\% & 100\,\% \\
\bottomrule
\end{tabular}
\end{adjustbox}
\end{table}

We observe consistent performance across all behaviors for Random Isolation Forests, One-Class SVM (OC-SVM), and Elliptic Envelope. However, Local Outlier Factor displays slightly lower Recall values, especially for ``Stopping'', indicating a potential lower sensitivity in identifying instances of this particular behavior:
\begin{itemize}
    \item {\em Cruising:} All classifiers perform well in recognizing the ``Cruising'' behavior, with Recall ranging from $88.8\%$ to $94.4\%$, indicating their ability to identify instances of this class. Precision is consistent at $100\%$, suggesting that when the classifiers identify the ``Cruising'' behavior, they are highly accurate.
    \item {\em Accelerating:} The classifiers excel in identifying instances of ``Accelerating'' with Recall values ranging from $90.3\%$ to $99.3\%$. Precision is consistently $100\%$, indicating that when the classifiers identify ``Accelerating'', they do so with high accuracy.
    \item {\em Stopping:} The performance varies for the ``Stopping'' action with Recall ranging from $38.6\%$ to $88.4\%$. Precision is consistent at $100\%$, implying high accuracy when the classifiers detect instances of ``Stopping''.
    \item {\em Steering Right and Steering Left:} The classifiers show strong performance in recognizing both ``Steering Right'' and ``Steering Left'' behaviors, with Recall values ranging from $92.0\%$ to $94.5\%$. Precision is consistent at $100\%$, indicating high accuracy in identifying these turning maneuvers.
\end{itemize}

In conclusion, the evaluation of one-class classifiers shows overall good performance.
Three out of four classifiers show an overall high performance, except local outlier factor.
For OC-SVM the recall is generally highest, except for the Stopping action, which is however one of the key actions to explain.
Following these results, we recommend the usage of random isolation forests as the one-class classifier, due to its consistently high, although not the highest, performance and better interpretability than elliptic envelope or OC-SVM.

\begin{table}[h]
\centering
\caption{Recall and Precision metrics for various multi-class classifiers}
\begin{adjustbox}{width=\textwidth,center}
\begin{tabular}{lccccccccccc}
\toprule
Classifier & \multicolumn{2}{c}{Cruising} & \multicolumn{2}{c}{Accelerate} & \multicolumn{2}{c}{Stop} & \multicolumn{2}{c}{Steering Right} & \multicolumn{2}{c}{Steering Left} \\
\cmidrule(r){2-3} \cmidrule(lr){4-5} \cmidrule(lr){6-7} \cmidrule(lr){8-9} \cmidrule(l){10-11}
& Recall & Precision & Recall & Precision & Recall & Precision & Recall & Precision & Recall & Precision \\
\midrule
Random Forest & 24.7\% & 1.6\% & 22.0\% & 50.5\% & 39.3\% & 27.1\% & 26.8\% & 43.8\% & 23.7\% & 0.9\% \\
Neural Network & 32.8\% & 1.8\% & 17.7\% & 51.6\% & 35.7\% & 26.0\% & 19.8\% & 45.4\% & 30.0\% & 0.8\% \\
Logistic Regression & 24.6\% & 1.1\% & 16.5\% & 46.8\% & 33.0\% & 18.3\% & 13.1\% & 41.5\% & 27.7\% & 0.9\% \\
Gradient Boosting & 28.3\% & 1.9\% & 16.3\% & 54.7\% & 42.7\% & 26.7\% & 32.1\% & 48.4\% & 25.8\% & 0.8\% \\
SVM & 3.0\% & 0.9\% & 74.3\% & 48.1\% & 13.3\% & 23.0\% & 5.9\% & 43.7\% & 10.3\% & 0.8\% \\
\bottomrule
\end{tabular}
\end{adjustbox}
\label{tab:classifier_metrics}
\end{table}

\subsubsection{Evaluation of Multi-Class Classifiers}
We further evaluate multi-class classifiers that do score an object-pair relation chain in terms of which action it has most likely caused.
Here, we consider Random Forest, Neural Network, Logistic Regression, Gradient Boosting, and Support Vector Machine (SVM) as classification models.
Results are shown in Table~\ref{tab:classifier_metrics}.
Here, we observe weaker and more mixed results than in the one-class classifier setting, even for the random forest model and SVM, which are conceptually similar to the models applied in the previous experiment.
This difference might stem from the conceptual switch between one-class classification where each classifier must only distinguish between the normal behavior and everything else, and the multi-class case where a precise action must be assigned to each behavior respectively object-pair relation chain.

One of the main challenges is the presence of noise in the data. Noise can result from various sources, such as through errors in the perception system of the ego vehicle or inaccuracies in labeling the actions. This noise can be redundant across different action labels, meaning that it does not provide useful information for distinguishing between the actions.

For instance, consider a scenario where the ego vehicle frequently encounters other vehicles that are stationary due to traffic congestion. The interactions with these stationary vehicles might be labeled with different actions such as "cruising", "stopping", or "steering right", depending on the ego vehicle's maneuver. 
However, the fact that the other vehicles are stationary is a piece of redundant information that is present across all these action labels. This redundant noise can make it more difficult for the multi-class classifiers to learn the distinguishing features of each action. Therefore, we notice a poor performance of the multi-class classifiers in Table~\ref{tab:classifier_metrics}. This phenomenon is not as strong when using one-class classifiers. Here, noise is less of an issue. This is because each classifier is trained to learn common observations on when a specific action is taken, but must not distinguish between the other actions, therefore allowing for a simpler decision boundary. 

\subsubsection{Cross-Region Validation}
To further assess the transferability of our approach, we divide the nuScenes dataset geographically into two distinct regions: Singapore (383 scenes) and Boston (467 scenes). One region is used for training and the other for testing and vice versa.
We thereby evaluate the adaptability of our model to the different geographical and infrastructural contexts. 
This regional split provides an interesting property: Besides the general differences in traffic behavior between regions, Singapore is left-hand traffic, whereas Boston is right-hand traffic. However, we expect that, due to the generalization to qualitative relations, the QXG will be capable of handling the transfer between the two regions, albeit with small performance decreases due to local peculiarities in traffic behavior and rules.

\begin{table}[ht]
\centering
\caption{One-Class Classifier Results Trained on Singapore and Validated on Boston\label{tab:singapore_boston}}
\begin{adjustbox}{width=\textwidth,center}
\begin{tabular}{lccccccccccc}
\toprule
Classifier & \multicolumn{2}{c}{Cruising} & \multicolumn{2}{c}{Accelerating} & \multicolumn{2}{c}{Stopping} & \multicolumn{2}{c}{Steering Right} & \multicolumn{2}{c}{Steering Left} \\
\cmidrule(r){2-3} \cmidrule(lr){4-5} \cmidrule(lr){6-7} \cmidrule(lr){8-9} \cmidrule(l){10-11}
& Recall & Precision & Recall & Precision & Recall & Precision & Recall & Precision & Recall & Precision \\
\midrule
Random Isolation Forests & 89.2\,\% & 100\,\% & 89.4\,\% & 100\,\% & 88.7\,\% & 100\,\% & 90.9\,\% & 100\,\% & 87.8\,\% & 100\,\% \\
One-Class SVM & 62.5\,\% & 100\,\% & 99.5\,\% & 100\,\% & 36.6\,\% & 100\,\% & 99.5\,\% & 100\,\% & 57.2\,\% & 100\,\% \\
Local Outlier Factor & 77.6\,\% & 100\,\% & 68.5\,\% & 100\,\% & 67.3\,\% & 100\,\% & 67.2\,\% & 100\,\% & 75.7\,\% & 100\,\% \\
Elliptic Envelope & 88.3\,\% & 100\,\% & 92.4\,\% & 100\,\% & 91.8\,\% & 100\,\% & 91.2\,\% & 100\,\% & 89.0\,\% & 100\,\% \\
\bottomrule
\end{tabular}
\end{adjustbox}
\end{table}

\begin{table}[ht]
\centering
\caption{One-Class Classifier Results Trained on Boston and Validated on Singapore\label{tab:boston_singapore}}
\begin{adjustbox}{width=\textwidth,center}
\begin{tabular}{lccccccccccc}
\toprule
Classifier & \multicolumn{2}{c}{Cruising} & \multicolumn{2}{c}{Accelerating} & \multicolumn{2}{c}{Stopping} & \multicolumn{2}{c}{Steering Right} & \multicolumn{2}{c}{Steering Left} \\
\cmidrule(r){2-3} \cmidrule(lr){4-5} \cmidrule(lr){6-7} \cmidrule(lr){8-9} \cmidrule(l){10-11}
& Recall & Precision & Recall & Precision & Recall & Precision & Recall & Precision & Recall & Precision \\
\midrule
Random Isolation Forests & 88.1\,\% & 100\,\% & 88.0\,\% & 100\,\% & 88.7\,\% & 100\,\% & 86.4\,\% & 100\,\% & 88.8\,\% & 100\,\% \\
One-Class SVM & 93.2\,\% & 100\,\% & 47.7\,\% & 100\,\% & 99.4\,\% & 100\,\% & 99.1\,\% & 100\,\% & 93.2\,\% & 100\,\% \\
Local Outlier Factor & 69.3\,\% & 100\,\% & 70.8\,\% & 100\,\% & 62.2\,\% & 100\,\% & 70.5\,\% & 100\,\% & 71.5\,\% & 100\,\% \\
Elliptic Envelope & 86.7\,\% & 100\,\% & 87.2\,\% & 100\,\% & 90.2\,\% & 100\,\% & 91.3\,\% & 100\,\% & 91.3\,\% & 100\,\% \\
\bottomrule
\end{tabular}
\end{adjustbox}
\end{table}

For the experiment, only one-class classifiers are trained, based on their superior performance in the earlier experiments.
Results are shown in Table~\ref{tab:singapore_boston} and \ref{tab:boston_singapore}.
The results are generally consistent with the earlier experiment where the training data followed the data split provided by nuScenes that contained scenes from both regions in the training and test sets.
We observe small decreases in the recall, which may be attributed to less diverse training data due to specific local traffic behavior in each region or small differences in traffic rules.
Still, the QXG representation is generic and transferable enough to allow correct action explanations for different regions even when major changes like switching from left-hand traffic to right-hand traffic occur.

\begin{itemize}
    \item {\em Cruising:} All classifiers perform well in recognizing the ``Cruising'' behavior. When trained on the Singapore data and validated on Boston data, Recall ranges from 62.5\% (OC-SVM) to 89.2\% (Random Isolation Forests). When trained on Boston data and validated on Singapore data, Recall ranges from 69.3\% (Local Outlier Factor) to 93.2\% (OC-SVM).
    \item {\em Accelerating:} The classifiers excel in identifying instances of ``Accelerating''. When trained on Singapore data and validated on Boston, Recall values range from 68.5\% (Local Outlier Factor) to 99.5\% (OC-SVM). When trained on Boston data and validated on Singapore, Recall ranges from 47.7\% (OC-SVM) to 88.0\% (Random Isolation Forests).
    \item {\em Stopping:} The performance varies for the ``Stopping'' action. When trained on Singapore data and validated on Boston, Recall ranges from 36.6\% (OC-SVM) to 91.8\% (Elliptic Envelope). When trained on Boston data and validated on Singapore, Recall ranges from 62.2\% (Local Outlier Factor) to 99.4\% (OC-SVM).
    \item {\em Steering Right and Steering Left:} The classifiers show strong performance in recognizing both ``Steering Right'' and ``Steering Left'' behaviors. When trained on Singapore data and validated on Boston, Recall values for ``Steering Right'' range from 67.2\% (Local Outlier Factor) to 99.5\% (OC-SVM), and for ``Steering Left'' range from 57.2\% (OC-SVM) to 89.0\% (Elliptic Envelope). When trained on Boston data and validated on Singapore, Recall values for ``Steering Right'' range from 70.5\% (Local Outlier Factor) to 99.1\% (OC-SVM), and for ``Steering Left'' range from 71.5\% (Local Outlier Factor) to 93.2\% (OC-SVM). 
\end{itemize}
As in the earlier experiment, precision is consistent at 100\% for all classifiers over all the actions.

\subsection{(RQ3) Scene Understanding Through Action Interpretation}

\subsubsection{Interpreting actions in safety-critical situations}
Of specific relevance is action explanation in safety-critical situations where a correct and timely response of any automated system is expected. 
Not only in the undesirable cases where an accident or misbehavior occurred, but also for validation reasons that the correct signals from the scene are identified to cause the correct action.

These safety-critical scenarios are not present in the rich available real-world datasets \cite{wang_deepaccident_2023}. 
We circumvent this shortcoming by specifically modeling safety-relevant scenarios for illustration and explanation.
Other recent work focuses on the generation of accident scenarios to test automated driving systems \cite{wang_deepaccident_2023}.
However, this work is not directly applicable to our setting, and we aim to identify illustrative scenarios to explain specific action behaviors rather than test real automated driving systems.

\begin{figure}[h]
\centering
\includegraphics[width=\textwidth]{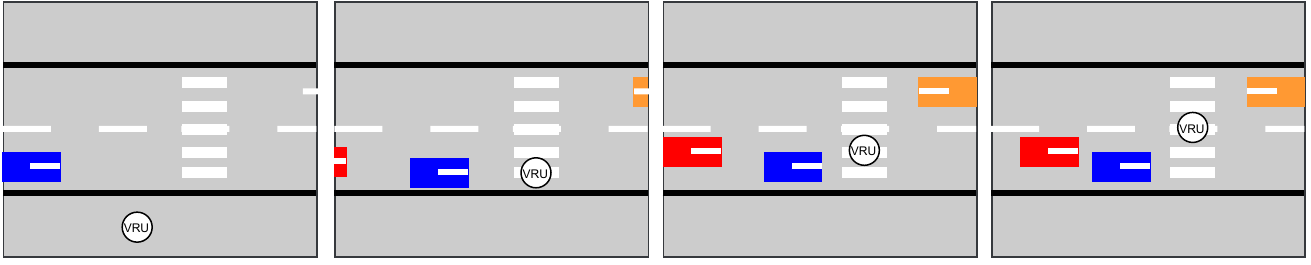}
\caption{Scenario of a pedestrian passing a cross-walk and causing stopping actions in the approaching vehicles.\label{fig:crossing}}
\end{figure}

For safety-critical situations, we illustrate two scenarios.
The first scenario, shown in Figure~\ref{fig:crossing}, describes an urban traffic situation where a pedestrian approaches and passes a crosswalk, which is likely to be anticipated by a driver or automated driving systems. 
At the same time, vehicles arrive on both lanes with one car (red) driving quickly towards the tail of another car (blue), indicating almost a takeover maneuver.
This is a typical everyday scenario that has clear rules to be followed and every involved participant has a clear expectation of what is supposed to be happening.

\begin{figure}[h]
\centering
\includegraphics[width=\textwidth]{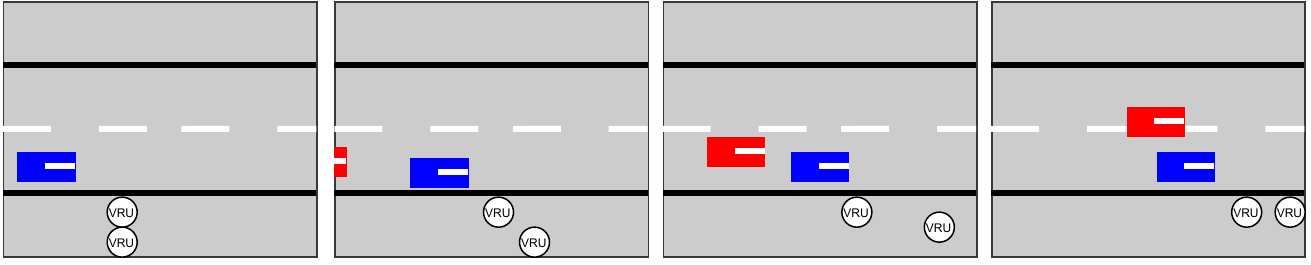}
\caption{Scenario of two pedestrians walking close to the roadway and causing stopping actions in the approaching blue vehicle while the red vehicle overtakes it\label{fig:closevru}}
\end{figure}

The second scenario, shown in Figure~\ref{fig:closevru}, depicts a potentially dangerous situation with an unclear future trajectory. 
Two pedestrians, e.g., a parent and a child, are on a sidewalk of the blue vehicle's lane. 
From the back, a red car approaches and aims to overtake the blue car.
One of the pedestrians suddenly moves quickly in the approximate direction of the lane, potentially entering or crossing it.
The second pedestrian follows it and finally can hinder it from entering the lane.
While this is a typical scenario involving unpredictable children, it is less clear what will be happening and how each involved participant will react.

We focus on the behavior of the blue vehicle as the main actor, even though the QXG is general enough to provide rationalizations for the actions of each object in the graph.
The most relevant action to be explained in both of these scenarios is the stopping action to avoid harm to the pedestrians.
However, there are contrary indications for alternatively choosing other actions.
For example, in the first scenario, the red car is approaching at high speed, which could potentially cause the blue car to cruise or even accelerate to maintain sufficient distance.

We describe both scenarios qualitatively as QXGs and confirm that the expected safe action, i.e., stopping, is correctly attributed to the pedestrian behavior in the scene using the random isolation forest action classifier trained with the nuScenes data.

\subsubsection{Interpreting actions in real-life scenarios}
\begin{figure}[th]
    \centering
    \includegraphics[width=\textwidth]{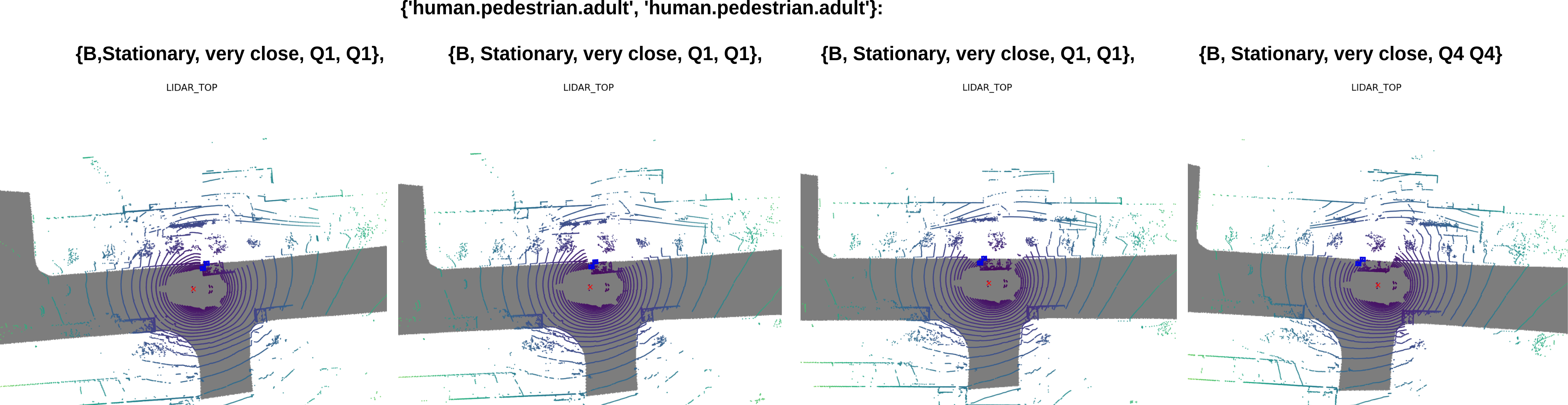}
\caption{Example action explanation of the ego car accelerating overlaid on the \lidar{} view: The ego car accelerates in proximity to two pedestrians on the left sidewalk, indicating a potentially hazardous situation. Order of relations: $RA$, $QTC_b$, $QDC$, $STAR_4$. The $B$ relation denotes spatial interaction, indicating an object situated inside another object.}
    \label{fig:exp_example1}
\end{figure}

\begin{figure}[th]
    \centering
    \includegraphics[width=\textwidth]{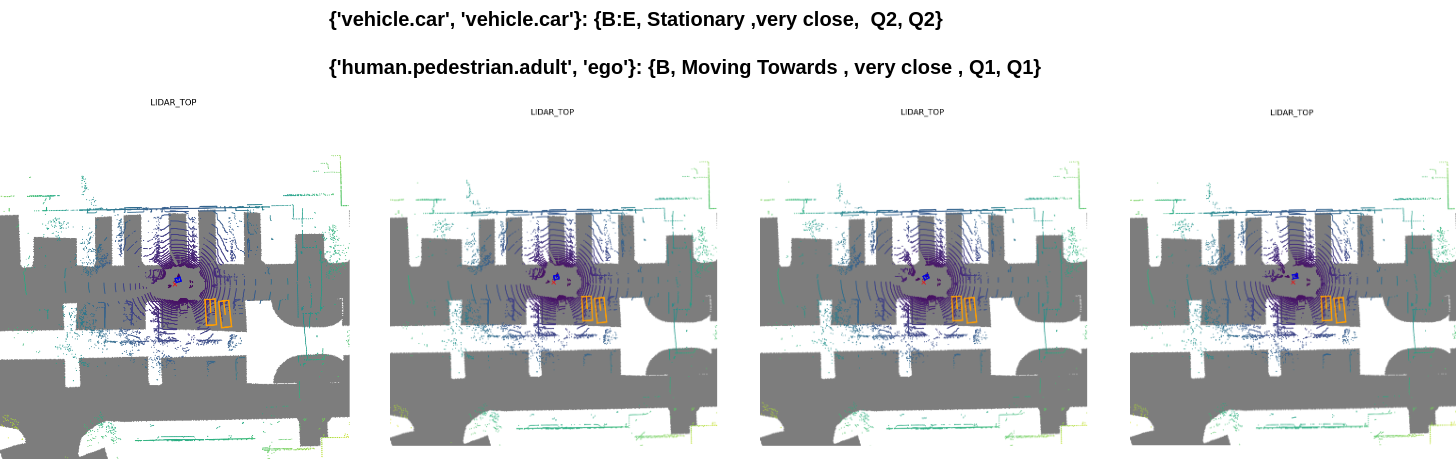}
 \caption{Example action explanation of the ego car coming to a stop overlaid on the \lidar{} view: The ego car stops, and two noteworthy scenarios unfold—one with a stationary pedestrian alarmingly close to the ego car and another with two parked cars unusually close to each other. Order of relations: $RA$, $QTC_b$, $QDC$, $STAR_4$. The $B$ relation indicates spatial interaction where an object is situated inside another object.}
    \label{fig:exp_example2}
\end{figure}

Finally, we illustrate two specific real-life scenarios taken from the nuScenes dataset: one scene involving the \textit{acceleration} of a vehicle (Figure~\ref{fig:exp_example1}) and the other one involving a \textit{stop} action (Figure~\ref{fig:exp_example2}).
By using one of our trained models, specifically Random Isolation Forests, we pinpoint the most significant object pairs in the scenes based on their scoring. 
In the first scenario (Figure~\ref{fig:exp_example1}), where the ego car accelerates, the following situation unfolds: two pedestrians remain stationary in very close proximity to the ego car, positioned on its left side. 
The classifier identifies these behavior patterns as deviating from typical expectations, and demanding careful consideration. 
This classification could function as an alert for the vehicle's occupants or serve as an explanation for any potentially anomalous actions taken by the ego car. Such insights play a vital role in advancing our comprehension of automated driving vehicle behavior and refining their decision-making algorithms.

In the second scenario (Figure~\ref{fig:exp_example2}), two noteworthy situations emerge: 
First, a pedestrian remains stationary, positioned within the bounding box of the ego car, presenting a spatial awareness concern for the automated driving system. 
Second, an intriguing arrangement features two parked cars positioned unusually close to each other. 
Similar to the previous example, these unusual behavioral patterns deviate from standard observations, indicating the potential need for a comprehensive analysis. 
Again, they could serve as critical alerts for the vehicle's occupants, drawing attention to potential deficiencies or anomalies in the vehicle's perception or decision-making system.

To summarize what is analyzed by these safety-critical and real-life scenarios, we find that QXG augmented with action explanations is relevant and useful to determine and interpret actions identified by our classifiers. As such, we answer RQ3 about the capability of QXG to interpret the ego car actions in real-life scenarios.

\section{Limitations and Challenges}
Our proposed approach is efficient in qualitatively representing scenes and explaining actions, however here are three main limitations/challenges. 
\begin{itemize}
    \item \textbf{Lack of Fine-Grained Action Representation:} The action explanation method currently uses a high-level representation of actions based on heuristics, such as "cruising", "accelerating", "stopping", "steering right", and "steering left". This coarseness is a limitation since there are degrees of maneuvers and complex actions that involve a combination of basic maneuvers not included in this representation.
    \item \textbf{Focus on Dynamic Objects:} The QXG primarily focuses on capturing interactions between dynamic objects. It does not take into account static objects or elements of the scene such as traffic lights, road signs, or roads, which can significantly influence the ego vehicle's actions. This could limit the method's ability to provide a complete explanation of the ego vehicle's actions in certain scenarios.
    \item \textbf{Dependence on Perception:} The proposed method relies on the perception collected from the sensors of the ego vehicle to identify and track objects in the scene. Therefore, any errors or limitations in the perception, such as difficulty in detecting small or distant objects, occluded objects, could affect the accuracy of the QXG-based action explanation. A main challenge would be to use qualitative reasoning techniques such as path consistency to overcome these issues.
\end{itemize}

\section{Conclusion}

Establishing a symbolic and qualitative comprehension of the vehicle's surroundings enhances communication not only with the car passengers, but also with other vehicles, vulnerable road users, and external auditors. 
It improves road safety, reliability, and trustworthiness. 

In this paper, we have introduced the Qualitative Explainable Graph (QXG), which is a spatio-temporal representation of driving scene that can be constructed incrementally in real-time.
It is versatile enough to not only permit the explanation of ego vehicle actions, but also to generalize to explanations for every acting object in a scene.
Through this QXG representation, we can provide action explanations for any object in the scene and from the perspective of any object in the scene. 
Beyond the usability for explanations in the ego vehicle, it can explain or anticipate the behavior of pedestrians or cyclists and support the traffic surveillance.
The key advantage of employing a qualitative scene representation lies in introspection and in-depth analysis capabilities. 
Action explanation is performed through action-specific action explainability classifiers that score object pairs and their spatio-temporal relations with regard to their likeliness to have caused an action.
We experimentally demonstrated the QXG construction and applicability of action explanations on a real-world dataset.

In future work, we will extend the application of QXGs for multi-vehicle scene understanding in connected mobility, facilitating intelligible explanations for road users, and exploring advanced message-passing techniques to enhance the action explanation process.
Using qualitative scene representation, we aim to contribute to the interoperability, safety, and reliability of automated driving systems in real-world environments. Establishing trust through transparent and comprehensible decision-making processes is a key requirement for societal acceptance and adoption.

\section*{Acknowledgment}
This work is funded by the European Commission through the AI4CCAM project (Trustworthy AI for Connected, Cooperative Automated Mobility) under grant agreement No 101076911, the TAILOR project under agreement No 952215, and by the AutoCSP project of the Research Council of Norway, grant number 324674. Experiments were performed on the Experimental Infrastructure for Exploration of Exascale Computing (eX3), which is financially supported by the Research Council of Norway under contract 270053.

\bibliographystyle{unsrtnat}

\bibliography{references_helge}

\begin{thebibliography}{28}
\providecommand{\natexlab}[1]{#1}
\providecommand{\url}[1]{\texttt{#1}}
\expandafter\ifx\csname urlstyle\endcsname\relax
  \providecommand{\doi}[1]{doi: #1}\else
  \providecommand{\doi}{doi: \begingroup \urlstyle{rm}\Url}\fi

\bibitem[Yang et~al.(2019)Yang, Wang, Liu, and Deng]{yang_scene_2019}
Shun Yang, Wenshuo Wang, Chang Liu, and Weiwen Deng.
\newblock Scene {Understanding} in {Deep} {Learning}-{Based} {End}-to-{End}
  {Controllers} for {Autonomous} {Vehicles}.
\newblock \emph{IEEE Transactions on Systems, Man, and Cybernetics: Systems},
  49\penalty0 (1):\penalty0 53--63, January 2019.
\newblock ISSN 2168-2232.
\newblock \doi{10.1109/TSMC.2018.2868372}.
\newblock URL \url{https://ieeexplore.ieee.org/document/8480450}.
\newblock Conference Name: IEEE Transactions on Systems, Man, and Cybernetics:
  Systems.

\bibitem[Muhammad et~al.(2022)Muhammad, Hussain, Ullah, Ser, Rezaei, Kumar,
  Hijji, Bellavista, and de~Albuquerque]{muhammad_vision-based_2022}
Khan Muhammad, Tanveer Hussain, Hayat Ullah, Javier~Del Ser, Mahdi Rezaei,
  Neeraj Kumar, Mohammad Hijji, Paolo Bellavista, and Victor Hugo~C.
  de~Albuquerque.
\newblock Vision-{Based} {Semantic} {Segmentation} in {Scene} {Understanding}
  for {Autonomous} {Driving}: {Recent} {Achievements}, {Challenges}, and
  {Outlooks}.
\newblock \emph{IEEE Transactions on Intelligent Transportation Systems},
  23\penalty0 (12):\penalty0 22694--22715, 2022.
\newblock \doi{10.1109/TITS.2022.3207665}.

\bibitem[Muhammad et~al.(2021)Muhammad, Ullah, Lloret, Ser, and
  de~Albuquerque]{muhammad_deep_2021}
Khan Muhammad, Amin Ullah, Jaime Lloret, Javier~Del Ser, and Victor Hugo~C.
  de~Albuquerque.
\newblock Deep {Learning} for {Safe} {Autonomous} {Driving}: {Current}
  {Challenges} and {Future} {Directions}.
\newblock \emph{IEEE Transactions on Intelligent Transportation Systems},
  22\penalty0 (7):\penalty0 4316--4336, 2021.
\newblock \doi{10.1109/TITS.2020.3032227}.

\bibitem[Omeiza et~al.(2022)Omeiza, Webb, Jirotka, and
  Kunze]{omeiza_explanations_2022}
Daniel Omeiza, Helena Webb, Marina Jirotka, and Lars Kunze.
\newblock Explanations in {Autonomous} {Driving}: {A} {Survey}.
\newblock \emph{IEEE Transactions on Intelligent Transportation Systems},
  23\penalty0 (8):\penalty0 10142--10162, 2022.
\newblock \doi{10.1109/TITS.2021.3122865}.

\bibitem[Nastjuk et~al.(2020)Nastjuk, Herrenkind, Marrone, Brendel, and
  Kolbe]{nastjuk_what_2020}
Ilja Nastjuk, Bernd Herrenkind, Mauricio Marrone, Alfred Brendel, and Lutz
  Kolbe.
\newblock What drives the acceptance of autonomous driving? {An} investigation
  of acceptance factors from an end-user's perspective.
\newblock \emph{Technological Forecasting and Social Change}, 161, October
  2020.
\newblock \doi{10.1016/j.techfore.2020.120319}.

\bibitem[Atakishiyev et~al.(2023)Atakishiyev, Salameh, Yao, and
  Goebel]{atakishiyev_explainable_2023}
Shahin Atakishiyev, Mohammad Salameh, Hengshuai Yao, and Randy Goebel.
\newblock Explainable {Artificial} {Intelligence} for {Autonomous} {Driving}:
  {A} {Comprehensive} {Overview} and {Field} {Guide} for {Future} {Research}
  {Directions}, February 2023.
\newblock URL \url{http://arxiv.org/abs/2112.11561}.
\newblock arXiv:2112.11561 [cs].

\bibitem[Kim et~al.(2021)Kim, Rohrbach, Akata, Moon, Misu, Chen, Darrell, and
  Canny]{kim_toward_2021}
Jinkyu Kim, Anna Rohrbach, Zeynep Akata, Suhong Moon, Teruhisa Misu, Yi-Ting
  Chen, Trevor Darrell, and John Canny.
\newblock Toward explainable and advisable model for self-driving cars.
\newblock \emph{Applied AI Letters}, 2\penalty0 (4):\penalty0 e56, 2021.
\newblock \doi{https://doi.org/10.1002/ail2.56}.
\newblock URL \url{https://onlinelibrary.wiley.com/doi/abs/10.1002/ail2.56}.

\bibitem[Caesar et~al.(2020)Caesar, Bankiti, Lang, Vora, Liong, Xu, Krishnan,
  Pan, Baldan, and Beijbom]{caesar_nuscenes_2020}
Holger Caesar, Varun Bankiti, Alex~H. Lang, Sourabh Vora, Venice~Erin Liong,
  Qiang Xu, Anush Krishnan, Yu~Pan, Giancarlo Baldan, and Oscar Beijbom.
\newblock {nuScenes}: {A} {Multimodal} {Dataset} for {Autonomous} {Driving}.
\newblock In \emph{Proceedings of the {IEEE}/{CVF} {Conference} on {Computer}
  {Vision} and {Pattern} {Recognition} ({CVPR})}, June 2020.

\bibitem[Dylla et~al.(2015)Dylla, Lee, Mossakowski, Schneider, Delden, Ven, and
  Wolter]{dylla_survey_2015}
Frank Dylla, Jae~Hee Lee, Till Mossakowski, Thomas Schneider, André~Van
  Delden, Jasper Van~De Ven, and Diedrich Wolter.
\newblock A {Survey} of {Qualitative} {Spatial} and {Temporal} {Calculi}:
  {Algebraic} and {Computational} {Properties}.
\newblock \emph{ACM Computing Surveys}, 50\penalty0 (1):\penalty0 7:1--7:39,
  2015.
\newblock ISSN 0360-0300.
\newblock \doi{10.1145/3038927}.
\newblock URL \url{https://doi.org/10.1145/3038927}.

\bibitem[Allen(1983)]{allen_maintaining_1983}
James~F. Allen.
\newblock Maintaining knowledge about temporal intervals.
\newblock \emph{Communications of the ACM}, 26\penalty0 (11):\penalty0
  832--843, November 1983.
\newblock ISSN 0001-0782.
\newblock \doi{10.1145/182.358434}.
\newblock URL \url{https://doi.org/10.1145/182.358434}.

\bibitem[Renz and Nebel(2007)]{renz_qualitative_2007}
Jochen Renz and Bernhard Nebel.
\newblock Qualitative {Spatial} {Reasoning} {Using} {Constraint} {Calculi}.
\newblock In \emph{Handbook of {Spatial} {Logics}}, pages 161--215. Springer
  Netherlands, Dordrecht, 2007.
\newblock ISBN 978-1-4020-5586-7 978-1-4020-5587-4.
\newblock \doi{10.1007/978-1-4020-5587-4_4}.
\newblock URL \url{http://link.springer.com/10.1007/978-1-4020-5587-4_4}.

\bibitem[Westhofen et~al.(2022)Westhofen, Neurohr, Butz, Scholtes, and
  Schuldes]{westhofen_using_2022}
Lukas Westhofen, Christian Neurohr, Martin Butz, Maike Scholtes, and Michael
  Schuldes.
\newblock Using {Ontologies} for the {Formalization} and {Recognition} of
  {Criticality} for {Automated} {Driving}, 2022.
\newblock \_eprint: 2205.01532.

\bibitem[Suchan et~al.(2021)Suchan, Bhatt, and
  Varadarajan]{suchan_commonsense_2021}
Jakob Suchan, Mehul Bhatt, and Srikrishna Varadarajan.
\newblock Commonsense visual sensemaking for autonomous driving – {On}
  generalised neurosymbolic online abduction integrating vision and semantics.
\newblock \emph{Artificial Intelligence}, 299:\penalty0 103522, 2021.
\newblock ISSN 0004-3702.
\newblock \doi{https://doi.org/10.1016/j.artint.2021.103522}.
\newblock URL
  \url{https://www.sciencedirect.com/science/article/pii/S0004370221000734}.

\bibitem[Xue et~al.(2018)Xue, Fang, and Zhang]{xue_survey_2018}
Jian-Ru Xue, Jian-Wu Fang, and Pu~Zhang.
\newblock A {Survey} of {Scene} {Understanding} by {Event} {Reasoning} in
  {Autonomous} {Driving}.
\newblock \emph{International Journal of Automation and Computing}, 15\penalty0
  (3):\penalty0 249--266, June 2018.
\newblock ISSN 1476-8186, 1751-8520.
\newblock \doi{10.1007/s11633-018-1126-y}.
\newblock URL \url{http://link.springer.com/10.1007/s11633-018-1126-y}.

\bibitem[Bessiere et~al.(2017)Bessiere, Koriche, Lazaar, and
  O'Sullivan]{bessiere_constraint_2017}
Christian Bessiere, Frédéric Koriche, Nadjib Lazaar, and Barry O'Sullivan.
\newblock Constraint acquisition.
\newblock \emph{Artificial Intelligence}, 244:\penalty0 315--342, March 2017.
\newblock ISSN 0004-3702.
\newblock \doi{10.1016/j.artint.2015.08.001}.
\newblock URL
  \url{https://www.sciencedirect.com/science/article/pii/S0004370215001162}.

\bibitem[Belaid et~al.(2022)Belaid, Belmecheri, Gotlieb, Lazaar, and
  Spieker]{belaid_geqca_2022}
Mohamed-Bachir Belaid, Nassim Belmecheri, Arnaud Gotlieb, Nadjib Lazaar, and
  Helge Spieker.
\newblock {GEQCA}: {Generic} {Qualitative} {Constraint} {Acquisition}.
\newblock In \emph{36th {AAAI} {Conference} {On} {Artificial} {Intelligence}},
  February 2022.

\bibitem[Campbell et~al.(2010)Campbell, Egerstedt, How, and
  Murray]{campbell_autonomous_2010}
Mark Campbell, Magnus Egerstedt, Jonathan~P. How, and Richard~M. Murray.
\newblock Autonomous driving in urban environments: approaches, lessons and
  challenges.
\newblock \emph{Philosophical Transactions: Mathematical, Physical and
  Engineering Sciences}, 368\penalty0 (1928):\penalty0 4649--4672, 2010.
\newblock ISSN 1364-503X.
\newblock URL \url{https://www.jstor.org/stable/20752685}.

\bibitem[Montemerlo et~al.(2009)Montemerlo, Becker, Bhat, Dahlkamp, Dolgov,
  Ettinger, Haehnel, Hilden, Hoffmann, Huhnke, Johnston, Klumpp, Langer,
  Levandowski, Levinson, Marcil, Orenstein, Paefgen, Penny, Petrovskaya,
  Pflueger, Stanek, Stavens, Vogt, and Thrun]{montemerlo_junior_2009}
Michael Montemerlo, Jan Becker, Suhrid Bhat, Hendrik Dahlkamp, Dmitri Dolgov,
  Scott Ettinger, Dirk Haehnel, Tim Hilden, Gabe Hoffmann, Burkhard Huhnke,
  Doug Johnston, Stefan Klumpp, Dirk Langer, Anthony Levandowski, Jesse
  Levinson, Julien Marcil, David Orenstein, Johannes Paefgen, Isaac Penny, Anna
  Petrovskaya, Mike Pflueger, Ganymed Stanek, David Stavens, Antone Vogt, and
  Sebastian Thrun.
\newblock Junior: {The} {Stanford} {Entry} in the {Urban} {Challenge}.
\newblock In Martin Buehler, Karl Iagnemma, and Sanjiv Singh, editors,
  \emph{The {DARPA} {Urban} {Challenge}: {Autonomous} {Vehicles} in {City}
  {Traffic}}, Springer {Tracts} in {Advanced} {Robotics}, pages 91--123.
  Springer, Berlin, Heidelberg, 2009.
\newblock ISBN 9783642039911.
\newblock \doi{10.1007/978-3-642-03991-1_3}.
\newblock URL \url{https://doi.org/10.1007/978-3-642-03991-1_3}.

\bibitem[Shladover(1995)]{shladover_review_1995}
Steven~E. Shladover.
\newblock Review of the {State} of {Development} of {Advanced} {Vehicle}
  {Control} {Systems} ({AVCS}).
\newblock \emph{Vehicle System Dynamics}, 24\penalty0 (6-7):\penalty0 551--595,
  July 1995.
\newblock ISSN 0042-3114, 1744-5159.
\newblock \doi{10.1080/00423119508969108}.
\newblock URL
  \url{http://www.tandfonline.com/doi/abs/10.1080/00423119508969108}.

\bibitem[Thrun(2010)]{thrun_toward_2010}
Sebastian Thrun.
\newblock Toward robotic cars.
\newblock \emph{Communications of the ACM}, 53\penalty0 (4):\penalty0 99--106,
  April 2010.
\newblock ISSN 0001-0782, 1557-7317.
\newblock \doi{10.1145/1721654.1721679}.
\newblock URL \url{https://dl.acm.org/doi/10.1145/1721654.1721679}.

\bibitem[Dolgov et~al.(2010)Dolgov, Thrun, Montemerlo, and
  Diebel]{dolgov_path_2010}
Dmitri Dolgov, Sebastian Thrun, Michael Montemerlo, and James Diebel.
\newblock Path {Planning} for {Autonomous} {Vehicles} in {Unknown}
  {Semi}-structured {Environments}.
\newblock \emph{The International Journal of Robotics Research}, 29\penalty0
  (5):\penalty0 485--501, April 2010.
\newblock ISSN 0278-3649, 1741-3176.
\newblock \doi{10.1177/0278364909359210}.
\newblock URL \url{http://journals.sagepub.com/doi/10.1177/0278364909359210}.

\bibitem[Ruoyu et~al.(2020)Ruoyu, Darui, Hang, Daihan, Ning, and
  Jianguang]{ruoyu_data-driven_2020}
Liu Ruoyu, Zhang Darui, Yang Hang, Wang Daihan, Bian Ning, and Zhou Jianguang.
\newblock A {Data}-{Driven} {Radar} {Object} {Detection} and {Clustering}
  {Method} {Aided} by {Camera}.
\newblock pages 2020--01--5035, February 2020.
\newblock \doi{10.4271/2020-01-5035}.
\newblock URL \url{https://www.sae.org/content/2020-01-5035/}.

\bibitem[Burgard et~al.(2008)Burgard, Brock, and
  Stachniss]{burgard_map-based_2008}
Wolfram Burgard, Oliver Brock, and Cyrill Stachniss.
\newblock Map-{Based} {Precision} {Vehicle} {Localization} in {Urban}
  {Environments}.
\newblock In \emph{Robotics: {Science} and {Systems} {III}}, pages 121--128.
  MIT Press, 2008.
\newblock ISBN 9780262255868.
\newblock URL \url{https://ieeexplore.ieee.org/document/6280129}.

\bibitem[Ou et~al.(2018)Ou, Bedawi, Koesdwiady, and Karray]{ou_predicting_2018}
Chaojie Ou, Safaa~Mahmoud Bedawi, Arief~B. Koesdwiady, and Fakhri Karray.
\newblock Predicting {Steering} {Actions} for {Self}-{Driving} {Cars} {Through}
  {Deep} {Learning}.
\newblock In \emph{2018 {IEEE} 88th {Vehicular} {Technology} {Conference}
  ({VTC}-{Fall})}, pages 1--5, August 2018.
\newblock \doi{10.1109/VTCFall.2018.8690657}.
\newblock URL \url{https://ieeexplore.ieee.org/abstract/document/8690657}.
\newblock ISSN: 2577-2465.

\bibitem[Xu et~al.(2017)Xu, Gao, Yu, and Darrell]{xu_end--end_2017}
Huazhe Xu, Yang Gao, Fisher Yu, and Trevor Darrell.
\newblock End-{To}-{End} {Learning} of {Driving} {Models} {From}
  {Large}-{Scale} {Video} {Datasets}.
\newblock pages 2174--2182, 2017.
\newblock URL
  \url{https://openaccess.thecvf.com/content_cvpr_2017/html/Xu_End-To-End_Learning_of_CVPR_2017_paper.html}.

\bibitem[Chi and Mu(2017)]{chi_learning_2017}
Lu~Chi and Yadong Mu.
\newblock Learning {End}-to-{End} {Autonomous} {Steering} {Model} from
  {Spatial} and {Temporal} {Visual} {Cues}.
\newblock In \emph{Proceedings of the {Workshop} on {Visual} {Analysis} in
  {Smart} and {Connected} {Communities}}, {VSCC} '17, pages 9--16, New York,
  NY, USA, October 2017. Association for Computing Machinery.
\newblock ISBN 9781450355063.
\newblock \doi{10.1145/3132734.3132737}.
\newblock URL \url{https://doi.org/10.1145/3132734.3132737}.

\bibitem[Pedregosa et~al.(2011)Pedregosa, Varoquaux, Gramfort, Michel, Thirion,
  Grisel, Blondel, Prettenhofer, Weiss, Dubourg, Vanderplas, Passos,
  Cournapeau, Brucher, Perrot, and Duchesnay]{pedregosa_scikit-learn_2011}
Fabian Pedregosa, Gaël Varoquaux, Alexandre Gramfort, Vincent Michel, Bertrand
  Thirion, Olivier Grisel, Mathieu Blondel, Peter Prettenhofer, Ron Weiss,
  Vincent Dubourg, Jake Vanderplas, Alexandre Passos, David Cournapeau,
  Matthieu Brucher, Matthieu Perrot, and Édouard Duchesnay.
\newblock Scikit-learn: {Machine} {Learning} in {Python}.
\newblock \emph{Journal of Machine Learning Research}, 12\penalty0
  (Oct):\penalty0 2825--2830, 2011.
\newblock ISSN ISSN 1533-7928.
\newblock URL \url{http://www.jmlr.org/papers/v12/pedregosa11a.html}.

\bibitem[Wang et~al.(2023)Wang, Kim, Ji, Xie, Ge, Chen, Li, and
  Luo]{wang_deepaccident_2023}
Tianqi Wang, Sukmin Kim, Wenxuan Ji, Enze Xie, Chongjian Ge, Junsong Chen,
  Zhenguo Li, and Ping Luo.
\newblock {DeepAccident}: {A} {Motion} and {Accident} {Prediction} {Benchmark}
  for {V2X} {Autonomous} {Driving}, August 2023.
\newblock URL \url{http://arxiv.org/abs/2304.01168}.
\newblock arXiv:2304.01168 [cs].

\end{thebibliography}

\end{document}